\documentclass{amsproc}
\usepackage{tikz}
\usepackage[margin=1in]{geometry}
\usepackage{tikz}
\usepackage{graphicx} 
\usepackage{multicol}
\usepackage{booktabs}
\usepackage{hyperref}
\usepackage{glossaries}
\usepackage{float}
\usepackage{expex}
\usepackage{setspace}
\usepackage{pgfplots}
\usepackage[utf8]{inputenc} 

\pgfplotsset{compat=1.18} 

\usepackage{multirow}

\usepackage{tikz}                
\usetikzlibrary{positioning,fit,arrows.meta,backgrounds, quotes, positioning, shapes.symbols}
\usetikzlibrary{arrows.meta,
                chains,
                positioning,
                shapes.geometric
                }

\usepackage{graphicx}
\usepackage{booktabs}
\usepackage{tikz}
\usetikzlibrary{trees}
\usetikzlibrary{shapes, positioning}

\usepackage{subcaption}
\usepackage{pgfplotstable}
\usepackage{arydshln}
\tikzset{
    module/.style={%
        draw, rounded corners,
        minimum width=#1,
        minimum height=7mm,
        font=\sffamily
        },
    module/.default=2cm,
    >=LaTeX
}
\tikzset{
    module1/.style={%
        draw, rounded corners,
        minimum width=220mm,
        align=center,
        minimum height=15mm,
        font=\sffamily
        },
    module/.default=2cm,
    >=LaTeX
}

\tikzset{
    module2/.style={%
        draw, rounded corners,
        minimum width=40mm,
        align=center,
        minimum height=15mm,
        font=\sffamily
        },
    module/.default=2cm,
    >=LaTeX
}

\tikzset{support/.style = {shape=signal, 
                 signal to=west and east,
                 font=\sffamily,
                 top color=white, bottom color=green!30}}

\tikzset{attack/.style = {shape=signal, 
                 signal to=west and east,
                 font=\sffamily,
                 top color=white, bottom color=red!30}}

\title{Cyborg Data: Merging Human with AI Generated Training Data}
\author{Kai North and Christopher Ormerod}
\date{July 2024}

\begin{document}

\begin{abstract}
Automated scoring (AS) systems used in large-scale assessment have traditionally used small statistical models that require a large quantity of hand-scored data to make accurate predictions, which can be time-consuming and costly. Generative Large Language Models are trained on many tasks and have shown impressive abilities to generalize to new tasks with little to no data. While these models require substantially more computational power to make predictions, they still require some fine-tuning to meet operational standards. Evidence suggests that these models can exceed human-human levels of agreement even when fine-tuned on small amounts of data. With this in mind, we propose a model distillation pipeline in which a large generative model, a Teacher, teaches a much smaller model, a Student. The Teacher, trained on a small subset of the training data, is used to provide scores on the remaining training data, which is then used to train the Student. We call the resulting dataset ``Cyborg Data", as it combines human and machine-scored responses.  Our findings show that Student models trained on ``Cyborg Data"  show performance comparable to training on the entire dataset, while only requiring 10\% of the original hand-scored data.


\end{abstract}

\maketitle

\begin{multicols}{2}

\section{Introduction}

Hand-scoring is time-consuming and costly to perform at scale. Automated essay scoring (AES) uses statistical models to assign grades that approximate hand-scoring \cite{ke_automated_2019}. Many states opt for AES as either a complete scoring solution or as part of a hybrid automated scoring system, which combines automated scoring with human scoring to improve the accuracy, efficiency, and consistency of the scoring process \cite{lottridge_psychometric_2023}. While the cost of AES in large-scale assessment is typically a fraction of what hand-scoring costs, the exact cost depends on many factors \cite{matta_cost_2022}. Large closed-source Generative Language Model (GLM)s such as GPT-4 \cite{openai_gpt-4_2023} have garnered significant interest in educational circles for their potential uses in automated scoring and content creation \cite{gimpel_unlocking_2023}. Instead of using these models directly to perform AES in large-scale assessment, this article concerns how these models can be used to teach smaller models to drive down the cost of AES development. 

Developing an AES model involves substantial upfront costs, with the creation of high-quality training data being the most resource-intensive component. The process requires hiring content experts who follow strict rubrics to evaluate essays. Depending on the quality control measures in place, each essay receives scores from at least two raters to measure scoring consistency, with additional reads sometimes employed for resolution. The scoring becomes more complex for trait-based assessment, where raters must evaluate and assign separate scores to multiple aspects of writing: technical elements like spelling and grammar (conventions), the quality of explanations and clarity (elaboration), and proper essay structure (organization) \cite{shermis_contrasting_2013, mathias_asap_2018}. Reducing the amount of training data required to meet operational standards \cite{williamson_framework_2012} could significantly lower the costs of AES model development.

Large closed-source GLMs such as GPT-4 \cite{openai_gpt-4_2023} have become particularly popular \cite{gimpel_unlocking_2023}. Attempts to leverage these models to perform essay scoring has had limited success \cite{xiao_human-ai_2024} when compared to traditional LLM-based approaches \cite{rodriguez_language_2019, uto_automated_2020}. Furthermore, since these closed-source models can only be accessed using APIs, their use in large-scale assessment is fraught with privacy, security, and environmental concerns \cite{bulut_rise_2024}. While closed-source models have received most of the attention, it is worth recognizing the impressive leaps and strides in performance that open-source GLMs have made. The various iterations of models, such as Phi-3 \cite{abdin_phi-3_2024}, Llama-3 \cite{aimeta_llama_2024}, Mistral \cite{jiang_mistral_2023}, Gemma \cite{gemma_team_gemma_2024}, and QWen \cite{yang_qwen2_2024}, can be fine-tuned directly or using parameter-efficient methods \cite{xu_parameter-efficient_2023}. Recent efforts using fine-tuned open-source GLMs \cite{ormerod_automated_2024_fixed} have demonstrated better performance than similar approaches using closed-source GLMs \cite{xiao_automation_2024}.

To perform AES at scale, there are many benefits to solutions that are computationally efficient \cite{ormerod_automated_2021, mayfield_should_2020}. While AES using transformer-based architectures is now well-established \cite{rodriguez_language_2019, uto_automated_2020}, more recent GLMs possess an order of magnitude more parameters. The size of these models, combined with extensive pretraining, enables zero-shot and few-shot learning capabilities \cite{brown_language_2020}. We propose that these more advanced models can serve as teachers to smaller models through a distillation process. Our approach uses a large GLM that, after training on just a small set of manually scored essays, can generate additional scored responses to expand the training dataset. In this framework, we call the larger model the ``Teacher" and the smaller model the ``Student", and the expanded dataset is called the ``augmented dataset" (otherwise referred to as the ``cyborg dataset").

This motivates the following research questions (RQs):

\begin{itemize}
\item[\textbf{RQ1}] Do Student models trained on augmented datasets achieve performance comparable to those trained on entirely human hand-scored data?
\item[\textbf{RQ2}] Do Student models exacerbate or introduce any additional bias compared to models trained exclusively on hand-scored data? 
\end{itemize}

To answer the above RQs, we experimented with utilizing scores generated by LLMs within an AES pipeline. 

This article is organized as follows: In \S \ref{sec:background}, we consider the background to this work, covering AES and model distillation more broadly. In \S \ref{sec:data}, we review the PERSUADE dataset, including the basic characteristics, the way we translate the data into prompts for scoring, and the synthetic data regime. In \S \ref{sec:method}, we detail the models and training techniques used to train the Teacher and Student models and how the final performance was evaluated against baselines. We follow this with a presentation of the results in \S \ref{sec:results}, which covers empirical performance gains by this pipeline and calculations relevant to bias. Finally, we discuss the implications of our results and relevant future work in \S \ref{sec:discussion}. 

\section{Background and Related Work}\label{sec:background}

AES research aims to streamline assessment and reduce costs for large-scale evaluations. Public datasets supporting this research include the Kaggle ASAP AES datasets \cite{shermis_contrasting_2013} and the PERSUADE corpus \cite{crossley_persuasive_2022}. Dominant approaches to provide AES have included frequency-based approaches with hand-crafted features \cite{attali_automated_2006, page_project_2003}, convolutional and recurrent neural networks \cite{dong_attention-based_2017, taghipour_neural_2016}, transformer-based encoder architectures \cite{rodriguez_language_2019, taghipour_neural_2016, uto_neural_2020}, and transformer-based generative approaches \cite{ormerod_automated_2024_fixed, xiao_human-ai_2024}. 

The transformer-based architecture is an important development \cite{vaswani_attention_2017}, however, the architecture alone is not the entire reason that these models have become successful. The success also hinges upon the architecture facilitating effective pretraining \cite{devlin_bert_2018, radford_improving_2018}. The cost to produce training data for AES is just a specific example of a more general principle: that labeled data is expensive to create. Pretraining utilizes a large corpus of unlabeled data in a semisupervised task. Examples include masked word prediction, demonstrated by the Bidirectional Encoder Representations by Transformers (BERT) model \cite{devlin_bert_2018} or next-word prediction, demonstrated by the Generative Pretrained Transformer (GPT) model \cite{radford_improving_2018}. Fine-tuning pretrained models requires significantly fewer training examples than training from scratch on downstream tasks, evident from GPT and BERT's performance on a range of benchmarks \cite{wang_glue_2019}. The effects of pretraining and data size was also explored in the context of AES \cite{ormerod_effects_2021}. 

When these models were introduced for AES \cite{rodriguez_language_2019}, they used an order of magnitude more parameters than methods that had been tried and tested before \cite{attali_automated_2006, dong_attention-based_2017, page_project_2003, taghipour_neural_2016}, prompting some to question the appropriateness of using these models for AES \cite{mayfield_should_2020}. Researchers in NLP addressed this issue more broadly by introducing models that improved the efficiency of the transformer architecture. Modifications such as bottlenecks \cite{sun_mobilebert_2020}, convolutional normalization layers \cite{jiang_convbert_2020}, and weight sharing \cite{lan_albert_2020} led to computationally more efficient models with fewer parameters while still delivering comparable performance. In parallel, researchers explored training efficiencies such as those employed to train the Efficiently Learning an Encoder that Classifies Token Replacements Accurately (ELECTRA) model \cite{clark_electra_2020}. The ELECTRA model, which was pretrained as a discriminator \cite{clark_electra_2020, he_deberta_2021}, performed AES as accurately as BERT \cite{ormerod_automated_2021}. This model has become a benchmark for efficient LLMs used in AES pipelines in large-scale assessment \cite{lottridge_psychometric_2023, ormerod_automated_2021}.

While many of these architectural efficiencies have not been incorporated into the latest generation of GLMs, one idea that has stood the test of time is model distillation \cite{schmidhuber_learning_1992}, which was applied to create DistilBERT \cite{sanh_distilbert_2020}. This process uses a larger model to teach a model with fewer parameters either directly or by using generated data. This method has been applied to GLMs such as Phi-3 \cite{abdin_phi-3_2024} and Deepseek's latest range of R1 models \cite{deepseek-ai_deepseek-r1_2025}. In these works, the smaller models perform better when trained, or fine-tuned, on a large corpus of synthetic data created by a larger model. 

This idea of fine-tuning on synthetic data can be very beneficial, especially in cases where the following may be true:
\begin{enumerate}
\item{The cost of obtaining a large corpus of high-quality data is very high.}
\item{The training data has too few instances of a particular class to be accurate.}
\end{enumerate}
One area where synthetic data has greatly improved the model accuracy has been Grammatical Error Correction (GEC). Researchers employed round-trip translation to build a substantial pre-training dataset, followed by fine-tuning on a smaller, high-quality corpus to enhance model performance \cite{stahlberg_synthetic_2021}. 

Synthetic data has been immensely useful when training requires large quantities of data. For image classification tasks using convolutional neural networks, researchers have utilized synthetic data derived from 3D models to improve the accuracy of classifying rare objects or scenarios \cite{tremblay_training_2018, rasmussen_development_2022}. These methods demonstrate how synthetic data can be leveraged to augment limited real-world datasets, potentially improving model performance and reducing annotation costs. 

The largest issue with synthetic data is the quality and fidelity of the data. If the data does not resemble the original data, or is too inaccurate, it can lead to model collapse \cite{shumailov_ai_2024}. For synthetic data to make sense, we need some assurances that the data is of a sufficiently high quality. Many have attempted to use GLMs to perform AES \cite{xiao_automation_2024, stahl_exploring_2024}, however, these efforts have not yielded agreements that are on par with the state-of-the-art \cite{xie_automated_2022}, or even a direct applications of BERT \cite{rodriguez_language_2019, uto_automated_2020}. Recently, researchers found that fine-tuning open-source GLMs using parameter efficient methods \cite{xu_parameter-efficient_2023} such as QLoRA \cite{dettmers_qlora_2023} to be much more effective than prompt-tuning larger closed-source models \cite{ormerod_automated_2024_fixed}. The key observation is that the pretraining and instruction tuning administered to GLMs provides an excellent starting point from which we can fine-tune a scoring model. In this pipeline, provided that the Teacher is more accurate than the smaller model trained on the same data, the Teacher can improve the results of the Student. 

\section{Data}\label{sec:data}

This study utilizes the PERSUADE corpus, which contains 25,996 essays annotated with argumentative components and relations, holistic essay scores, and a range of demographic information, including data on race, gender, and student disability \cite{crossley_persuasive_2022}. The essays were authored by students from a diverse background representative of the US population from grade 6 to 12. The minimum length is 150 words, and each essay was a response to one of 15 unique persuasive writing prompts.

\begin{table}[H]
\begin{center}
\begin{tabular}{l r r r | r } \toprule
& & & & {\bf Avg.} \\
{\bf Grade} & {\bf Train} & {\bf Test} & {\bf Total} & {\bf Len.}\\ \midrule
6 & 688   & 684  & 1372 & 294.6 \\
8 & 5614  & 4015 & 9629 & 374.9\\
9 & 1831  & 235  & 2066 & 426.6 \\
10 & 4654 & 3620 & 8274 & 407.6\\
11 & 1863 & 1220 & 3083 & 610.9\\
12 & 243  & 161  & 404  & 469.0\\
Unk. & 701 & 467 & 1168 & 452.5\\ \midrule
Total & 15594 & 10402 & 25996 & 418.1 \\ \bottomrule
\end{tabular}
\end{center}
\caption{Descriptive statistics of the training and test split for the PERSUADE corpus. \label{tab:data_desciption}}
\end{table}

\begin{table}[H]
\begin{center}
\begin{tabular}{l l r r} \toprule
key & Student group & Train & Test \\ \midrule
WC & White/Caucasian & 7012 & 4559\\
HL &Hispanic/Latino & 3869 & 2691\\
BA & Black/African American & 2975 & 1984\\
AP & Asian/Pacific Islander & 1072 & 671 \\
TW & Two or more & 598 & 424\\
NT & American Indian/Alaskan & 68 & 73\\ \midrule
ELL & English Language Learner & 1330 &  914 \\
DE & Disadvantaged Economically & 5391 & 4252 \\
ID & Identified Disability & 1516 & 1172 \\ \bottomrule
\end{tabular}
\end{center}
\caption{The main student group populations in the train and test set. \label{tab:subgroups}}
\end{table}

The PERSUADE corpus was used as part of the Feedback Prize competition hosted by Kaggle for the automatic identification of persuasive elements \cite{crossley_persuasive_2022}. As such, the dataset contains a pre-defined train and test split amounting to 15,594 essays for training and 10,402 essays for testing (Table \ref{tab:data_desciption}). We used this same split to train models to predict holistic essays scores.

\begin{figure}[H]
\centering
\scalebox{0.9}{
\begin{tikzpicture}[node distance=4cm]
    \tikzstyle{block} = [very thick, rectangle, draw, text width=0.8\linewidth, rounded corners=10pt, minimum height=6em];
    
    \node [block] (rect1) {
        \textbf{\underline{User:}} After reading the essay, assign a holistic score based on the rubric below. For the following evaluations you will need to use a grading scale between 1 (minimum) and 6 (maximum).
        
        \vspace{1em}
        \texttt{\{essay\}} \\
        \vspace{1em}
        \texttt{\{rubric\}} \\
        \vspace{1em}
        \textbf{\underline{Assistant:}} 

        \hspace{1cm} - Score:  \texttt{\{score\}} 
    };
\end{tikzpicture}
}
\caption{The template used to create the prompt for training the Teacher GLMs.}
\label{fig:prompt1}
\end{figure}

\subsection{Prompt Template}

Training modern GLMs is performed in three stages: pertaining, instruction tuning, and refinement \cite{openai_gpt-4_2023}. During pertaining, the model is trained to predict the next token from a large corpus in a similar manner to the original GPT model \cite{radford_improving_2018}. The instruction tuning phase seeks to train the model to perform tasks, while the refinement stage uses reinforcement learning to make the outputs more useful and amenable. During both the instruction tuning and refinement process, the models are subjected to instructions that adhere to a particular format \cite{ouyang_training_2022}. 

To train the model to perform scoring, we constructed a prompt that adheres to the same format as an instruction, followed by an assistant's response that provides the score. By experimenting with the model directly, we constructed a prompt that creates a score that is easy to parse and is based on the same instructions provided to human scorers. This was achieved by including a modified version of the grading rubric \footnote{Grading Rubric: Both source-independent and source-dependent rubrics available at github.com/scrosseye/persuade\_corpus\_2.0} inside the instructions alongside each student essay. The modifications of the original rubric included the omission of any tasks not related to the assignment of a holistic score as well as some slight rewording of the rubric by the LLM to be more easily understood. The prompt we constructed for training is shown in Figure \ref{fig:prompt1}.


\subsection{Synthetic Data}
The GLM (Teacher GLM) returned a holistic score in the specified range between 1 and 6 when provided with the prompt shown in Figure \ref{fig:prompt1}. We fine-tuned our Teacher GLM on different amounts of the original training data to obtain more accurate holistic score predictions. For example, we trained our GLM nine times with each training iteration using 10\% more of the original training data. The test set used during these training iterations consisted of the remaining unused samples of the original train set. In other words, training iteration one was trained on 10\% (1,559 instances) and predicted 90\% (14,035 instances) of the original train set, whereas training iteration two was trained on 20\% (3,119 instances) and predicted 80\% (12,475 instances) of the original train set, and so on. We then used the predictions from each training iteration as synthetic data which we combined with that training iteration's train split to create an augmented dataset for each iteration.

By having multiple augmented datasets, we aimed to discover the minimum amount of original training data needed to obtain the highest quality synthetic predictions. In doing so, we wanted to uncover whether synthetic predictions could be used in place of original data to achieve similar state-of-the-art performance for AES, hence answering our first RQ.

\section{Method}\label{sec:method}

\subsection{Model Training}

There are two classes of models we need to train; there are models used to generate synthetic scores, which are large GLMs, and smaller models that are trained on the outputs of the larger models. As discussed above, we call these models the Teacher and the Student. 

\begin{table}[H]
\begin{tabular}{l l l r } \toprule 
    Role & Model & Ref. & \# Params. \\ \midrule
    Teacher & Llama 3.1 & \cite{aimeta_llama_2024} & 8.1B \\ \midrule
    Student & ModernBERT & \cite{warner_smarter_2024} & 160M\\
     & ELECTRA & \cite{clark_electra_2020} & 11M\\ \bottomrule
\end{tabular}
\caption{A basic description of the Teacher and Student Models used in this study.}
\end{table}

\vspace{.3cm}

\noindent {\bf The Teacher}: Our ``Teacher model" is based on the Llama 3.1 architecture \cite{aimeta_llama_2024}. This variant has been fine-tuned using instruction-based training and contains approximately 8 billion parameters. The model utilizes a decoder-only transformer structure with 32 layers. Each attention layer has 4,096 dimensions, complemented by a feed-forward layer of 14,336 dimensions. The model's vocabulary consists of 128,000 tokens, and it can natively process input sequences of up to 8,192 tokens in length.

The availability of the model weights allows us to run the model locally \footnote{https://huggingface.co/meta-llama/Meta-Llama-3-8B-Instruct}. This is particularly important in educational settings where the security of student data is an important factor \cite{bulut_rise_2024}. To maximize the accuracy of these models on the specific task on our modest hardware, we are required to explore parameter-efficient methods \cite{xu_parameter-efficient_2023}. In particular, we fine-tune this model using the Quantized Low-Rank adapters (QLoRa) method demonstrated by Dettmers et al. \cite{dettmers_qlora_2023}. The QLoRa method was recently used to demonstrate the ability of GLMs to perform AES and provide feedback \cite{ormerod_automated_2024_fixed}. 

At the core of the Llama architecture \cite{touvron_llama_2023}, and more transformers more generally, is the attention mechanism \cite{vaswani_attention_2017}. The attention mechanism relies on multiple linear layers, taking the canonical form
\begin{equation}\label{eq:linear}
L(x) = Wx + b.
\end{equation}
The LoRa method replaces \eqref{eq:linear} with the operator
\begin{eqnarray}
\tilde{L}(x) =  (W+\Delta) x,
\Delta = BA
\end{eqnarray}
If $W \in \mathrm{Mat}_{m\times n}$ then $B \in \mathrm{Mat}_{m\times k}$ and $A \in \mathrm{Mat}_{k\times n}$ for some $k$. The multiplication, $\Delta = BA$ is a convenient parameterization of an element of $\mathrm{Mat}_{m\times n}$ of a rank that is less than or equal to $k$, which is assumed to be less than $m$, hence the term Low-Rank adaptation. In this method, the matrix $W$ is frozen; we are only applying updates to the matrices $A$ and $B$ using the memory efficient 8-bit paged weighted Adam optimizer \cite{dettmers_8-bit_2022}. While LoRa is a general technique applied to any of the linear layers of a model, it is commonly applied to the linear layers associated with the attention mechanisms, i.e., the normalization layers applied to the inputs to obtain keys, queries, and values in attention. 

The instruction tuning phase of the model training for the Llama models includes a large number of diverse tasks, in a similar manner to the T5 model \cite{raffel_exploring_2020}. More generally, GLMs that have been instruction-tuned seem to exhibit a degree of transfer learning \cite{wang_superglue_2020}. It stands to reason that the three respective tasks associated with the AAE corpus may benefit from transfer learning. This means that, instead of training a model for each task, transfer learning facilitates the use of a single model for all three tasks. We do this simply by combining prompts associated with component identification, component classification, and stand identification into one dataset. 

\vspace{.3cm}

\noindent {\bf The Student}: The first model chosen was the ELECTRA (Efficiently Learning an Encoder that Classifies Token Replacements Accurately) model, which uses a pre-training approach for a language model that offers an alternative to the masked language modeling used in BERT \cite{devlin_bert_2018}. Instead of masking tokens and asking the model to predict the original words, ELECTRA uses a generator-discriminator architecture where a small generator network proposes replacements for tokens in the text, and the main model (the discriminator) learns to distinguish between original and replaced tokens \cite{clark_electra_2020}. This training regime is more efficient than BERT's masked language modeling because it learns from all input tokens rather than just the masked ones, and it achieves better performance on downstream tasks with the same compute budget. The model's success demonstrates that detecting whether tokens have been replaced is an effective pre-training task for language understanding. We used a small version of this model with only 13 million parameters, which has been used in AES pipelines \cite{lottridge_psychometric_2023}.

Despite the success of the ELECTRA model in AES \cite{ormerod_automated_2021}, the architecture of ELECTRA model is design to only accept a maximum of only 512 tokens as input. ModernBERT is an encoder model that incorporates many of the architectural improvements that have been developed with the latest wave of GLMs \cite{warner_smarter_2024}. One of these developments is the use of Rotary Positional Embeddings (RoPE) \cite{su_roformer_2024}, which is a necessary component for context length extensions \cite{fu_data_2024}. ModerBERT uses pretraining using a context length of 512, which is then extended to 8196 by using RoPE layers with alternating rotary values and additional training. This model is perhaps a more appropriate benchmark as it is capable of taking the full essay as input, it was trained on an order of magnitude more tokens, and it represents more current architectural paradigms for transformer-based models \cite{xiong_layer_2020, shazeer_glu_2020}. 

Both Student models were trained using the cross-entropy loss function that compared the predicted scores with the true scores on the training set, using the Adam optimizer with Weight decay with a learning rate of $5\times 10^{-5}$ and a batch size of 4. A linear learning rate scheduler was applied over 10 epochs of training, and to simplify the training regime, no development set was used. This means that no early stopping mechanism was applied. When we compared the Student models to models trained only on the smaller subset of data the Teacher GLMs were trained on, we also used 10 epochs, despite the difference in the total number of training steps.

\begin{table*}
\centering
\begin{tabular}{c|rrrr | r r r r} \toprule
                                      & \multicolumn{4}{c}{QWK} & \multicolumn{4}{c}{SMD}  \\
    Original/     & \multicolumn{2}{c}{ELECTRA} & \multicolumn{2}{c}{Modern-BERT}  & \multicolumn{2}{c}{ELECTRA} & \multicolumn{2}{c}{Modern-BERT}\\ 
     Synthetic & Orig. & w/Aug. & Orig. & w/Aug. &  Orig. & w/Aug. & Orig. & w/Aug. \\  \midrule
      10\%/90\% &  0.658  & 0.809 &  0.799 & 0.817 & -0.079 & -0.078 & 0.081  & 0.165 \\
      20\%/80\% &  0.788  & 0.811 &  0.820 & 0.822 & -0.076 & -0.070 & 0.017  & 0.174\\
      30\%/70\% &  0.800  & 0.817 &  0.831 & 0.837 & -0.110 & -0.115 & 0.019  & 0.092\\
      40\%/60\% &  0.804  & 0.817 &  0.838 & 0.839 & -0.132 & -0.113 & 0.008  & 0.065\\
      50\%/50\% &  0.813  & 0.826 &  0.840 & 0.842 & -0.111 & -0.113 & -0.004 & 0.039\\
      60\%/40\% &  0.812  & 0.823 &  0.842 & 0.843 & -0.139 & -0.132 & -0.011 & 0.037\\
      70\%/30\% &  0.814  & 0.825 & 0.842 & 0.843 & -0.176 & -0.113 & -0.008 & 0.024\\
      80\%/20\% &  0.819  & 0.823 & 0.843 & 0.844 & -0.139 & -0.149 & -0.016 & 0.009\\
      90\%/10\% &  0.819  & 0.826 & 0.844 & 0.844 & -0.176 & -0.144 & -0.016 & 0.002\\
   
      \hdashline

     100\%/0\% & 0.813 & -  & 0.844  & - & - & - & - & -  \\ \bottomrule
   
\end{tabular}
\vspace{.3cm}
 \caption{Average QWKs across five experiments for holistic score prediction using varying numbers of original (Org.) and augmented training 
data (W/Aug.).}\label{results_lcp}

\end{table*}

\subsection{Validation and Metrics}

The Teacher and Student models were assessed using Quadratic Weighted Kappa (QWK) and Standardized Mean Difference (SMD), two critical metrics employed to evaluate the performance of AES for practical applications \cite{williamson_framework_2012}. Additionally, exact agreement or accuracy is considered as another key indicator of AES quality; however, this specific measure was not reported due to a lack of information on the exact match between the two raters in the original manuscript \cite{crossley_persuasive_2022}.

The weighted kappa is calculated by the ratio of the weighted sum of the products of observed instances ($O_{ij}$) and the weighted sum of their corresponding expected values ($E_{ij}$), where $i$ represents the score assigned by the first rater, $j$ represents the score assigned by the second rater,
\begin{equation}\label{eq:qwk}
\kappa = 1 - \frac{\sum w_{ij} O_{ij}}{\sum w_{ij} E_{ij}},
\end{equation}
which becomes quadratic weighted kappa when the quadratic weighting, defined by
\begin{equation}\label{eq:weight}
w_{ij} = \dfrac{(i-j)^2}{(N+1)^2},
\end{equation}
is applied. The SMD is defined by
\begin{equation}\label{eq:smd}
SMD(y_t, y_p) = \frac{\overline{y_p} - \overline{y_t}}{\sqrt{(\sigma(y_p)^2 + \sigma(y_t)^2)/2}},
\end{equation}
where we have used the notation $\overline{y}$ to denote the average.

To validate our approach, we conducted a straightforward experiment using the metrics described above. We define:
\begin{itemize}
\item $X$: The complete set of training data.
\item $Y$: The test set.
\item $M$: A pretrained model
\item $U \subset X$: a subset of the scored training data.
\end{itemize}
We created $X^U$, an augmented dataset formed by using a Teacher model (trained on $U$) to score the remaining data. 

Our success criterion is simple: if $M_V(Y)$ represented the expected model performance (measured by QWK) on the test set $Y$ when trained some set $V$, then we aim to demonstrate that
\[
M_{X^U}(Y) > M_U(Y).
\]
\begin{figure*}
\centering

\begin{subfigure}[b]{0.45\textwidth}
\centering
\begin{tikzpicture}[xscale=8, yscale=16]
\draw[very thick] (0.1,0.6) -- (0.1,.87);
\draw[blue, very thick] (0.1,0.8086180794492988) -- (0.2,0.8111037473434166);
\draw[blue, very thick] (0.2,0.8111037473434166) -- (0.3,0.8170765000300376);
\draw[blue, very thick] (0.3,0.8170765000300376) -- (0.4,0.8170186248997295);
\draw[blue, very thick] (0.4,0.8170186248997295) -- (0.5,0.8262008205758494);
\draw[blue, very thick] (0.5,0.8262008205758494) -- (0.6,0.8228903392728932);
\draw[blue, very thick] (0.6,0.8228903392728932) -- (0.7,0.8248019414764904);
\draw[blue, very thick] (0.7,0.8248019414764904) -- (0.8,0.8227274464485645);
\draw[blue, very thick] (0.8,0.8227274464485645) -- (0.9,0.8256280675581991);
\draw[cyan,very thick] (0.1,0.6576111122032671) -- (0.2,0.7872813379077692);
\draw[cyan,very thick] (0.2,0.7872813379077692) -- (0.3,0.800427678649047);
\draw[cyan,very thick] (0.3,0.800427678649047) -- (0.4,0.8039732344486108);
\draw[cyan,very thick] (0.4,0.8039732344486108) -- (0.5,0.8126719884086884);
\draw[cyan,very thick] (0.5,0.8126719884086884) -- (0.6,0.8124334412876968);
\draw[cyan,very thick] (0.6,0.8124334412876968) -- (0.7,0.8137748647095157);
\draw[cyan,very thick] (0.7,0.8137748647095157) -- (0.8,0.818534026171536);
\draw[cyan,very thick] (0.8,0.818534026171536) -- (0.9,0.8190697447516619);
\draw[red,very thick] (0.1,0.7985287969940327) -- (0.2,0.8195702823800708);
\draw[red,very thick] (0.2,0.8195702823800708) -- (0.3,0.8306290142457573);
\draw[red,very thick] (0.3,0.8306290142457573) -- (0.4,0.8381866646892334);
\draw[red,very thick] (0.4,0.8381866646892334) -- (0.5,0.8399947333440032);
\draw[red,very thick] (0.5,0.8399947333440032) -- (0.6,0.8420130812355977);
\draw[red,very thick] (0.6,0.8420130812355977) -- (0.7,0.8423129198879387);
\draw[red,very thick] (0.7,0.8423129198879387) -- (0.8,0.843167962102725);
\draw[red,very thick] (0.8,0.843167962102725) -- (0.9,0.8444637891396873);
\draw[brown,very thick] (0.1,0.8165513731825923) -- (0.2,0.8215957451179514);
\draw[brown,very thick] (0.2,0.8215957451179514) -- (0.3,0.8366282616000719);
\draw[brown,very thick] (0.3,0.8366282616000719) -- (0.4,0.8393338614607403);
\draw[brown,very thick] (0.4,0.8393338614607403) -- (0.5,0.8424853733204379);
\draw[brown,very thick] (0.5,0.8424853733204379) -- (0.6,0.8433541012350612);
\draw[brown,very thick] (0.6,0.8433541012350612) -- (0.7,0.8431999896307424);
\draw[brown,very thick] (0.7,0.8431999896307424) -- (0.8,0.8440040629137819);
\draw[brown,very thick] (0.8,0.8440040629137819) -- (0.9,0.8439094102416529);

\node at (0.1,0.57) {10\%};
\draw (0.1,.59) -- (0.1,.61);
\node at (0.2,0.57) {20\%};
\draw (0.2,.59) -- (0.2,.61);
\node at (0.3,0.57) {30\%};
\draw (0.3,.59) -- (0.3,.61);
\node at (0.4,0.57) {40\%};
\draw (0.4,.59) -- (0.4,.61);
\node at (0.5,0.57) {50\%};
\draw (0.5,.59) -- (0.5,.61);
\node at (0.6,0.57) {60\%};
\draw (0.6,.59) -- (0.6,.61);
\node at (0.7,0.57) {70\%};
\draw (0.7,.59) -- (0.7,.61);
\node at (0.8,0.57) {80\%};
\draw (0.8,.59) -- (0.8,.61);
\node at (0.9,0.57) {90\%};
\draw (0.9,.59) -- (0.9,.61);
\draw[dashed, black!50] (0.1, 0.65) -- (0.9,0.65);
\draw[dashed, black!50] (0.1, 0.7) -- (0.9,0.7);
\draw[dashed, black!50] (0.1, 0.75) -- (0.9,0.75);
\draw[dashed, black!50] (0.1, 0.8) -- (0.9,0.8);
\draw[dashed, black!50] (0.1, 0.85) -- (0.9,0.85);
\node at (0.1-0.04, 0.6) {0.6};
\node at (0.1-0.04, 0.65) {0.65};
\node at (0.1-0.04, 0.7) {0.7};
\node at (0.1-0.04, 0.75) {0.75};
\node at (0.1-0.04, 0.8) {0.8};
\node at (0.1-0.04, 0.85) {0.85};
\draw[very thick] (0.1,0.6) -- (0.9,0.6);
\node at (0.1-0.12, .73) {\rotatebox{90}{QWK}};
\end{tikzpicture}

\hspace{0.25cm} Percentage of Original Data Used
\end{subfigure}
\hfill
\begin{subfigure}[b]{0.45\textwidth}
\centering
\begin{tikzpicture}[xscale=8, yscale=12]

	\draw[very thick] (0.1,-0.2) -- (0.1,.2);
	\draw[blue, very thick] (0.1,-0.078089659819846) -- (0.2,-0.0698452244988664);
	\draw[blue, very thick] (0.2,-0.0698452244988664) -- (0.3,-0.11451081247045594);
	\draw[blue, very thick] (0.3,-0.11451081247045594) -- (0.4,-0.11272894871964939);
	\draw[blue, very thick] (0.4,-0.11272894871964939) -- (0.5,-0.11308256674209136);
	\draw[blue, very thick] (0.5,-0.11308256674209136) -- (0.6,-0.13161671314644174);
	\draw[blue, very thick] (0.6,-0.13161671314644174) -- (0.7,-0.11264161375179566);
	\draw[blue, very thick] (0.7,-0.11264161375179566) -- (0.8,-0.14939722643079295);
	\draw[blue, very thick] (0.8,-0.14939722643079295) -- (0.9,-0.14421253930093658);
	\draw[cyan,very thick] (0.1,-0.0793880953512598) -- (0.2,-0.0759201979301698);
	\draw[cyan,very thick] (0.2,-0.0759201979301698) -- (0.3,-0.11028139537293156);
	\draw[cyan,very thick] (0.3,-0.11028139537293156) -- (0.4,-0.132416090437097);
	\draw[cyan,very thick] (0.4,-0.132416090437097) -- (0.5,-0.11071806830751962);
	\draw[cyan,very thick] (0.5,-0.11071806830751962) -- (0.6,-0.138646050117251);
	\draw[cyan,very thick] (0.6,-0.138646050117251) -- (0.7,-0.17629320861209208);
	\draw[cyan,very thick] (0.7,-0.17629320861209208) -- (0.8,-0.13902359087110464);
	\draw[cyan,very thick] (0.8,-0.13902359087110464) -- (0.9,-0.1761046004978335);
	\draw[red,very thick] (0.1,0.08061603110609168) -- (0.2,0.017460639257859668);
	\draw[red,very thick] (0.2,0.017460639257859668) -- (0.3,0.018906522079300932);
	\draw[red,very thick] (0.3,0.018906522079300932) -- (0.4,0.007787619613087898);
	\draw[red,very thick] (0.4,0.007787619613087898) -- (0.5,-0.004291890994972433);
	\draw[red,very thick] (0.5,-0.004291890994972433) -- (0.6,-0.011355171654990235);
	\draw[red,very thick] (0.6,-0.011355171654990235) -- (0.7,-0.008369304756656933);
	\draw[red,very thick] (0.7,-0.008369304756656933) -- (0.8,-0.015819050339447433);
	\draw[red,very thick] (0.8,-0.015819050339447433) -- (0.9,-0.0157658083856922);
	\draw[brown,very thick] (0.1,0.1653405999211517) -- (0.2,0.17354578367758547);
	\draw[brown,very thick] (0.2,0.17354578367758547) -- (0.3,0.09228841804150824);
	\draw[brown,very thick] (0.3,0.09228841804150824) -- (0.4,0.0649935714729231);
	\draw[brown,very thick] (0.4,0.0649935714729231) -- (0.5,0.03936888463151947);
	\draw[brown,very thick] (0.5,0.03936888463151947) -- (0.6,0.03720739702638464);
	\draw[brown,very thick] (0.6,0.03720739702638464) -- (0.7,0.024424193109586368);
	\draw[brown,very thick] (0.7,0.024424193109586368) -- (0.8,0.008992755993849166);
	\draw[brown,very thick] (0.8,0.008992755993849166) -- (0.9,0.0015828959316658668);
\node at (0.1,-0.23) {10\%};
\draw[dashed, black!50] (0.1,-.2) -- (0.1,.2);
\node at (0.2,-0.23) {20\%};
\draw[dashed, black!50] (0.2,-.2) -- (0.2,.2);
\node at (0.3,-0.23) {30\%};
\draw[dashed, black!50] (0.3,-.2) -- (0.3,.2);
\node at (0.4,-0.23) {40\%};
\draw[dashed, black!50] (0.4,-.2) -- (0.4,.2);
\node at (0.5,-0.23) {50\%};
\draw[dashed, black!50] (0.5,-.2) -- (0.5,.2);
\node at (0.6,-0.23) {60\%};
\draw[dashed, black!50] (0.6,-.2) -- (0.6,.2);
\node at (0.7,-0.23) {70\%};
\draw[dashed, black!50] (0.7,-.2) -- (0.7,.2);
\node at (0.8,-0.23) {80\%};
\draw[dashed, black!50] (0.8,-.2) -- (0.8,.2);
\node at (0.9,-0.23) {90\%};
\draw[dashed, black!50] (0.9,-.2) -- (0.9,.2);
\draw[dashed, black!50] (0.1, -0.2) -- (0.9,-0.2);
\draw[dashed, black!50] (0.1, -0.15) -- (0.9,-0.15);
\draw[dashed, black!50] (0.1, -0.1) -- (0.9,-0.1);
\draw[dashed, black!50] (0.1, -0.05) -- (0.9,-0.05);
\draw[dashed, black!50] (0.1, 0.0) -- (0.9,0.0);
\draw[dashed, black!50] (0.1, 0.05) -- (0.9,0.05);
\draw[dashed, black!50] (0.1, 0.1) -- (0.9,0.1);
\draw[dashed, black!50] (0.1, 0.15) -- (0.9,0.15);
\node at (.1-0.05, -0.2) {-0.2};
\node at (.1-0.05, -0.15) {-0.15};
\node at (.1-0.05, -0.1) {-0.1};
\node at (.1-0.05, -0.05) {-0.05};
\node at (.1-0.05, 0.0) {0.0};
\node at (.1-0.05, 0.05) {0.05};
\node at (.1-0.05, 0.1) {0.1};
\node at (.1-0.05, 0.15) {0.15};
	\draw[very thick] (0.1,0.0) -- (0.9,0.0);

\node at (0.1-0.12, 0) {\rotatebox{90}{SMD}};

\end{tikzpicture}

\hspace{0.25cm} Percentage of Original Data Used

\end{subfigure}

\vspace{0.25cm}

\begin{tabular}{c}
\textcolor{red}{\rule{0.3cm}{0.2mm}} MBERT Orig. \hspace{0.5cm}
\textcolor{brown}{\rule{0.3cm}{0.2mm}} MBERT w/Aug. \hspace{0.5cm}
\textcolor{cyan}{\rule{0.3cm}{0.2mm}} ELECTRA Orig. \hspace{0.5cm}
\textcolor{blue}{\rule{0.3cm}{0.2mm}} ELECTRA w/Aug.
\end{tabular}

\caption{The average QWKs and SMDs for the ModernBERT and ELECTRA Apprentice models used for scoring when trained on original only (Orig.) and a percentage of the original train set plus remaining synthetic (w/Aug.), i.e. 10\% original and 90\% synthetic, 20\% original and 80\% synthetic and so on.} \label{merged1_2}
\end{figure*}
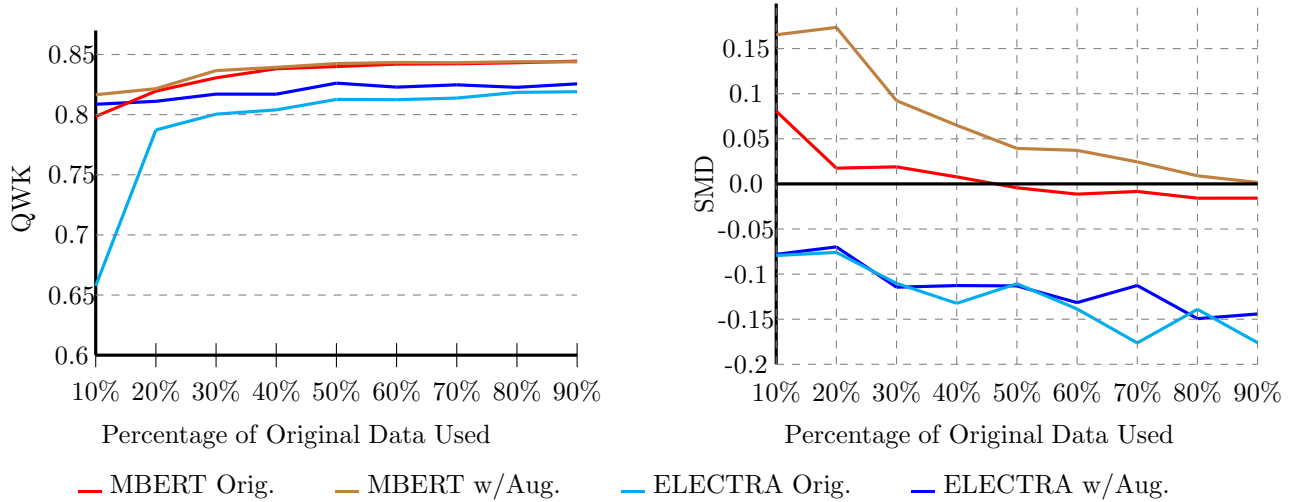

If we can demonstrate that this inequality holds, then we have shown that models trained on our augmented dataset $X^U$ outperform those trained solely on the original subset $U$ - a fair comparison since both scenarios assume we only have scores provided by $U$. For robust evaluation, we calculated the performance at each training set percentage by averaging results across five different randomly selected subsets $U$ of the specified size.

\section{Results}\label{sec:results}




\begin{table*}
\centering
\begin{tabular}{c|rrrr | r r r r} \toprule
                                      & \multicolumn{4}{c}{QWK} & \multicolumn{4}{c}{SMD}  \\
    Original/     & \multicolumn{2}{c}{ELECTRA} & \multicolumn{2}{c}{Modern-BERT}  & \multicolumn{2}{c}{ELECTRA} & \multicolumn{2}{c}{Modern-BERT}\\ 
     Synthetic & t-test &ks-test & t-test &ks-test &  t-test &ks-test & t-test &ks-test \\  \midrule
      10\%/90\% &  \textbf{0.01}& \textbf{0.008}& 0.537& 0.079& 0.556 & 0.357 & 0.395  & 0.079 \\
      20\%/80\% &  0.165& 0.079& 0.925& 0.079& 0.29 & 0.873 & 0.057  & 0.079 \\
      30\%/70\% &  \textbf{0.042}& 0.079& 0.53& 0.229& 0.3 & 0.873 & 0.096  & 0.079 \\
      40\%/60\% &  0.123& 0.229& 0.839& 0.079& 0.192 & 0.229 & 0.196  & 0.229 \\
      50\%/50\% &  \textbf{0.003}& \textbf{0.008}& 0.142& 0.873& 0.409 & 0.873 & \textbf{0.005}  & 0.079 \\
      60\%/40\% &  \textbf{0.023}& 0.079& 0.447& 0.357& 0.309 & 0.873 & \textbf{0.001}  & \textbf{0.008} \\
      70\%/30\% &  \textbf{0.003}& \textbf{0.008}& 0.726& 0.873& 0.076 & 0.079 & \textbf{0.004}  & \textbf{0.008} \\
      80\%/20\% &  0.286& 0.873& 0.798& 0.873& 0.276 & 0.357 & \textbf{0.024}  & \textbf{0.008} \\
      90\%/10\% &  \textbf{0.039}& 0.357& 0.628& 0.873& 0.131 & 0.079 & \textbf{0.009}  & 0.079 \\
   
\end{tabular}
\vspace{.3cm}
 \caption{P-values obtained from a standard pairwise t-test and ks-test between the QWKs and SMDs of the original compared to w/Aug. training sets across five experiments. Those $\displaystyle \leq$ 0.05 shown in bold.}\label{results_pvalues}
\end{table*}

It was discovered that our augmented datasets produced by our Teacher GLM improved the performance of our ELECTRA Student model at predicting student essay scores across all five experiments, regardless of the quantity of synthetic data used or original/synthetic training split. Table \ref{results_lcp} and Figure \ref{merged1_2} show the average QWK between the test set's gold essay scores and the essays scores predicted by our ELECTRA model for all five experiments when trained on differing quantities of original data compared to original data plus synthetic (w/Aug). ELECTRA achieved an average QWK of 0.658 when only trained on 10\% of the original training set (Org.). ELECTRA achieved a significantly greater average QWK of 0.809 (with a p-value $\displaystyle \leq$ 0.05; Table \ref{results_pvalues}) when trained on 10\% of the original training set plus the remaining 90\% of the training set as synthetic predictions (w/Aug.); our Teacher GLM generated these synthetic predictions by also having been trained on the same 10\% of the original training set as the Student model. The remaining augmented datasets with differing original/synthetic training splits also improved ELECTRA's overall performance, with performances on average being greater than those trained solely on the original training set. However, the difference in performance decreases with the more original data incorporated within the augmented dataset.

Our ModernBERT Student model also showed an increase in performance when being trained on an augmented training set containing 10\% of the original training set and the remaining 90\% as synthetic predictions provided by our Teacher GLM. ModernBERT achieved an average QWK of 0.799 across the five experiments when trained on only 10\% of the original training set (Org.). In contrast, ModernBERT achieved an average QWK of 0.817 when being trained on 10\%/90\% original and synthetic split (w/Aug.). This noticeable increase in performance is parallel to that observed by our ELECTRA Student model. However, unlike ELECTRA, our ModernBERT Student model showed less improved overall performance when incorporating more of the original data within the augmented training split. This may be a result of our Student ModernBERT model no longer being able to benefit from the synthetic predictions of our Teacher GLM, since ModernBERT has already achieved its highest possible on the given test set. Despite this, the performance of our augmented 10\% original and 90\% synthetic training set across both Student models and all experiments would suggest that state-of-the-art performances can be achieved with far fewer hand-annotated essay scores. In fact, for both our ELECTRA and ModernBERT Student models, only a difference of 0.004 and 0.027 in QWK respectively exists between using 10\% of the original training set plus synthetic compared to using 100\% of the original training set. This means that the performance achieved by our Student models when using 10\% of the original training set plus synthetic is almost identical to that of using the entire original training set.

Both Student models were also found to score each student essay more highly when being trained on more than 50\% of original data with remaining instances being synthetic from our Teacher GLM.  On average, ELECTRA and ModernBERT produced SMDs +0.012 and +0.06 higher respectively when being trained on a combination of original and synthetic data than compared to being trained on original data only (Figure \ref{merged1_2}). ModernBERT in particular assigned far higher scores to each essay when being trained on more synthetic data. For instance, ModernBERT achieved an average SMD of 0.174 when trained on 20\%  of the original training set plus the remaining 80\% as synthetic predictions. This SMD is  +0.157 greater than compare to when being trained on 80\% of the original training set only. The synthetic data produced by our Teacher GLM therefore has the potential to result in more favorable overall scoring. The following section examines whether these higher scores are the result of any form of bias to a particular demographic.

\begin{figure*}
    \centering

\begin{subfigure}{0.45\textwidth}
\begin{tikzpicture}
    \begin{axis}[
        ybar,
        symbolic x coords={10\%, 20\%, 30\%, 40\%, 50\%, 60\%, 70\%, 80\%, 90\%},
        xtick=data,
        ymin=-0.2,
        ymax=0.1,
        ylabel={SMD},
        xlabel={Percentage of Original Data Used},
        title={Gender Bias},
        legend style={at={(0.5,-0.3)}, anchor=north, draw=none, legend columns=1},
        legend image post style={xscale=0.8}, 
        legend cell align={left},
        width=8.5cm,
        height=6cm,
        bar width=4pt,
        cycle list name=color list,
        ytick align=inside,
        grid=both, 
        grid style={ymajor grid/.style={solid}, xmajor grid/.style={dashed, gray}}, 
        enlarge x limits=0.06
    ]

    \addplot[fill=blue!60, draw=blue!80] coordinates {(10\%, -0.116163354) (20\%, -0.125289604) (30\%, -0.07917079) (40\%, -0.072589334) (50\%, -0.06231011) (60\%, -0.05026104) (70\%, -0.039805672) (80\%, -0.03192729) (90\%, -0.017823372)};
            
    \addplot[fill=red!60, draw=red!80] coordinates {(10\%, -0.111409438) (20\%, -0.121182724) (30\%, -0.078749378) (40\%, -0.06879249) (50\%, -0.05627513) (60\%, -0.043149918) (70\%, -0.037667996) (80\%, -0.028036624) (90\%, -0.013678698)};

        \legend{Male, Female}

    \end{axis}
\end{tikzpicture}
\end{subfigure} 
\hfill
\begin{subfigure}{0.45\textwidth}
\begin{tikzpicture}
    \begin{axis}[
        ybar,
        symbolic x coords={10\%, 20\%, 30\%, 40\%, 50\%, 60\%, 70\%, 80\%, 90\%},
        xtick=data,
        ymin=-0.2,
        ymax=0.1,
        xlabel={Percentage of Original Data Used},
        title={ELL Bias},
        legend style={at={(0.5,-0.3)}, anchor=north, draw=none, legend columns=1},
        legend image post style={xscale=0.8}, 
        legend cell align={left},
        width=8.5cm,
        height=6cm,
        bar width=4pt,
        cycle list name=color list,
        ytick align=inside,
        yticklabels={,,}, 
        grid=both, 
        grid style={ymajor grid/.style={solid}, xmajor grid/.style={dashed, gray}}, 
        enlarge x limits=0.06
    ]

    \addplot[fill=black!60, draw=black!80] coordinates {(10\%, -0.11724847) (20\%, -0.123838534) (30\%, -0.080019914) (40\%, -0.072466934) (50\%, -0.061190552) (60\%, -0.046942818) (70\%, -0.038161656) (80\%, -0.029421112) (90\%, -0.016897648)};
            
    \addplot[fill=purple!60, draw=purple!80] coordinates {(10\%, -0.092489554) (20\%, -0.165445696) (30\%, -0.09632567) (40\%, -0.08130338) (50\%, -0.068670726) (60\%, -0.066550186) (70\%, -0.067077774) (80\%, -0.054938412) (90\%, -0.010160744)};
        
        \legend{Native English Speaker, English Language Learner}

    \end{axis}
\end{tikzpicture}
\end{subfigure}

\vspace{1em} 

\begin{subfigure}{0.45\textwidth}
\begin{tikzpicture}
    \begin{axis}[
        ybar,
        symbolic x coords={10\%, 20\%, 30\%, 40\%, 50\%, 60\%, 70\%, 80\%, 90\%},
        xtick=data,
        ymin=-0.2,
        ymax=0.1,
        ylabel={SMD},
        xlabel={Percentage of Original Data Used},
        title={Disability Bias},
        legend style={at={(0.5,-0.3)}, anchor=north, draw=none, legend columns=1},
        legend image post style={xscale=0.8}, 
        legend cell align={left},
        width=8.5cm,
        height=6cm,
        bar width=4pt,
        cycle list name=color list,
        ytick align=inside,
        grid=both, 
        grid style={ymajor grid/.style={solid}, xmajor grid/.style={dashed, gray}}, 
        enlarge x limits=0.06
    ]

    \addplot[fill=cyan!60, draw=cyan!80] coordinates {(10\%, -0.131914066) (20\%, -0.136148764) (30\%, -0.086543754) (40\%, -0.07349908) (50\%, -0.057999422) (60\%, -0.046103202) (70\%, -0.043720956) (80\%,-0.034457928) (90\%, -0.018640942)};
            
    \addplot[fill=orange!60, draw=orange!80] coordinates {(10\%, -0.123140978) (20\%, -0.150774156) (30\%, -0.076497012) (40\%, -0.067955772) (50\%, -0.066672686) (60\%, -0.05327434) (70\%, -0.056787674) (80\%, -0.044362292) (90\%, -0.018628022)};
    
        \legend{No Disability, With Disability}

    \end{axis}
\end{tikzpicture}
\end{subfigure}
\hfill
\begin{subfigure}{0.45\textwidth}
\begin{tikzpicture}
    \begin{axis}[
        ybar,
        symbolic x coords={10\%, 20\%, 30\%, 40\%, 50\%, 60\%, 70\%, 80\%, 90\%},
        xtick=data,
        ymin=-0.2,
        ymax=0.1,
        xlabel={Percentage of Original Data Used},
        title={Economic Status Bias},
        legend style={at={(0.5,-0.3)}, anchor=north, draw=none, legend columns=1},
        legend image post style={xscale=0.8}, 
        legend cell align={left},
        width=8.5cm,
        height=6cm,
        bar width=4pt,
        cycle list name=color list,
        ytick align=inside,
        yticklabels={,,}, 
        grid=both, 
        grid style={ymajor grid/.style={solid}, xmajor grid/.style={dashed, gray}}, 
        enlarge x limits=0.06
    ]

    \addplot[fill=cyan!60, draw=cyan!80] coordinates {(10\%, -0.13219032) (20\%, -0.12974076) (30\%, -0.080961312) (40\%, -0.071623508) (50\%, -0.056745438) (60\%, -0.045971822) (70\%, -0.045623458) (80\%, -0.036265186) (90\%, -0.020213234)};
            
    \addplot[fill=green!60, draw=green!80] coordinates {(10\%, -0.135377166) (20\%, -0.156255148) (30\%, -0.09573032) (40\%, -0.077892188) (50\%, -0.065283458) (60\%, -0.050887694) (70\%, -0.047218736) (80\%, -0.036667712) (90\%, -0.017557642)};

    \legend{No Economic Disadvantage, Economically Disadvantaged}

    \end{axis}
\end{tikzpicture}
\end{subfigure}

\caption{SMDs for the augmented dataset's essay scores for gender, English Language Learner (ELL) status, disability and economic status compared to their original scores provided by the PERSUADE corpus. Percentages are in relation to the amount of original data used within each training set. The remaining percentage being synthetic.} \label{bias_merge}
\end{figure*}
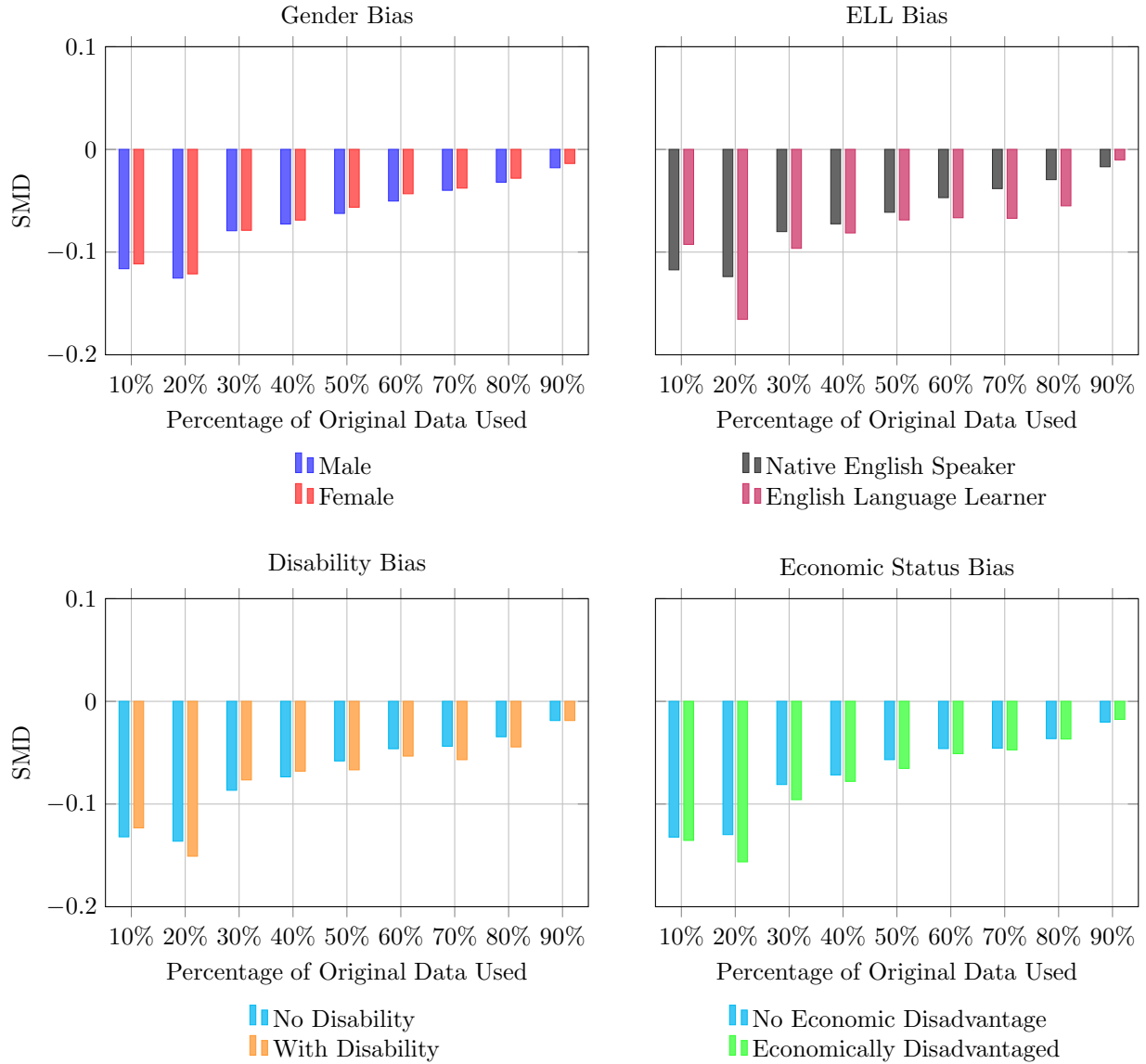
\begin{figure*}
\begin{tikzpicture}
    \begin{axis}[
        ybar,
        symbolic x coords={10\%, 20\%, 30\%, 40\%, 50\%, 60\%, 70\%, 80\%, 90\%},
        xtick=data,
        ymin=-0.2,
        ymax=0.1,
        ylabel={SMD},
        xlabel={Percentage of Original Data Used},
        title={Bias for Different Racial/Ethnic Groups},
        legend style={at={(0.5,-0.3)}, anchor=north, draw=none, legend columns=3},
        legend image post style={xscale=0.8}, 
        legend cell align={left},
        width=16cm,
        height=6cm,
        bar width=4pt,
        cycle list name=color list,
        ytick align=inside,
        grid=both, 
        grid style={ymajor grid/.style={solid}, xmajor grid/.style={dashed, gray}}, 
        enlarge x limits=0.06 
    ]
        \addplot[fill=cyan!60, draw=cyan!80] coordinates {(10\%, -0.10931736) (20\%, -0.121153716) (30\%, -0.071260334) (40\%, -0.063040974) (50\%, -0.055460246) (60\%, -0.039003856) (70\%, -0.037109774) (80\%, -0.026351392) (90\%, -0.013824394)};

        \addplot[fill=green!60, draw=green!80] coordinates {(10\%, -0.107145508) (20\%, -0.133570586) (30\%, -0.081007342) (40\%, -0.071519014) (50\%, -0.062036788) (60\%, -0.050746422) (70\%, -0.040629612) (80\%, -0.032529568) (90\%, -0.014093758)};

        \addplot[fill=blue!60, draw=blue!80] coordinates {(10\%, -0.147646408) (20\%, -0.18354871) (30\%, -0.169629416) (40\%, -0.127358498) (50\%, -0.109580964) (60\%, -0.086350778) (70\%, -0.05402799) (80\%, -0.037852984) (90\%, -0.035209492)};

        \addplot[fill=red!60, draw=red!80] coordinates {(10\%, -0.095614538) (20\%, -0.094120964) (30\%, -0.065119492) (40\%, -0.062501654) (50\%, -0.052200788) (60\%, -0.042387888) (70\%, -0.040428684) (80\%, -0.037338182) (90\%, -0.013584932)};

        \addplot[fill=orange!70, draw=orange!90] coordinates {(10\%, -0.12573198) (20\%, -0.128257856) (30\%, -0.087021502) (40\%, -0.07720581) (50\%, -0.062982458) (60\%, -0.050119786) (70\%, -0.039226992) (80\%, -0.029439312) (90\%, -0.017696918)};
        
        \addplot[fill=purple!60, draw=purple!80] coordinates {(10\%, -0.107702046) (20\%, -0.105961476) (30\%, -0.056131746) (40\%, -0.060735852) (50\%, -0.045712406) (60\%, -0.038233164) (70\%, -0.036730816) (80\%, -0.031941268) (90\%, -0.019186436)};
    
        \legend{Black/African American, Hispanic/Latino,  American Indian/Alaskan Native, Asian/Pacific Islander, White, Two or more races/Other}

    \end{axis}
\end{tikzpicture}
\caption{SMDs for the augmented dataset's essay scores for various racial/ethnic groups compared to their original scores provided by the PERSUADE corpus. Percentages are in relation to the amount of original data used within each training set. The remaining percentage being synthetic.}\label{bias_race}
\end{figure*}
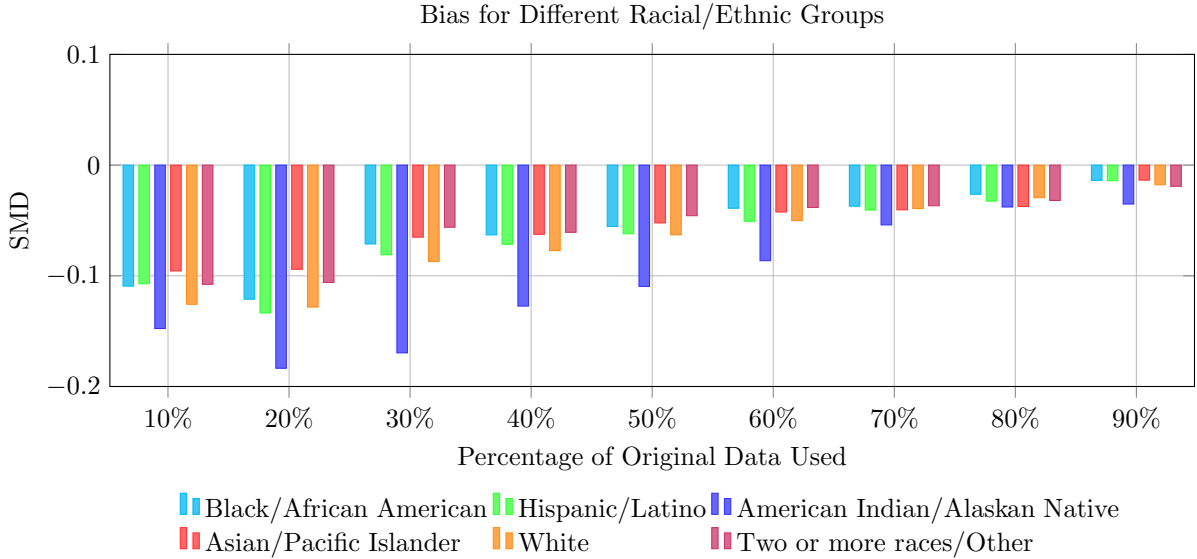

\begin{table*}
\begin{tabular}{lrrrrrrrrrr}
\toprule
& \multicolumn{9}{c}{Percentage of original training data used w/Aug data}& \\
Demographic  & 10\% & 20\% & 30\% & 40\% & 50\% & 60\% & 70\% & 80\% & 90\% & 100\% \\
\midrule
Male & 3.17 & 3.14 & 3.18 & 3.19 & 3.19 & 3.19 & 3.21 & 3.22 & 3.23 & 3.24 \\
Female & 3.41 & 3.41 & 3.43 & 3.43 & 3.43 & 3.43 & 3.45 & 3.45 & 3.46 & 3.47 \\
\midrule

Native English Speaker & 3.33 & 3.32 & 3.35 & 3.35 & 3.35 & 3.36 & 3.38 & 3.38 & 3.39 & 3.4 \\
English Language Learner & 2.75 & 2.63 & 2.71 & 2.67 & 2.66 & 2.68 & 2.69 & 2.68 & 2.71 & 2.72 \\

\midrule
No Disability & 3.39 & 3.4 & 3.43 & 3.44 & 3.44 & 3.44 & 3.46 & 3.46 & 3.48 & 3.48  \\
With Disability & 2.82 & 2.74 & 2.81 & 2.8 & 2.79 & 2.8 & 2.82 & 2.82 & 2.83 & 2.84  \\
\midrule
No Economic Disadvantage & 3.58 & 3.62 & 3.63 & 3.64 & 3.65 & 3.64 & 3.67 & 3.67 & 3.68 & 3.69\\
Economically Disadvantaged & 2.99 & 2.96 & 3.01 & 3.01 & 3.0 & 3.01 & 3.03 & 3.03 & 3.04 & 3.05 \\
\midrule
Black/African American & 3.11 & 3.06 & 3.1 & 3.13 & 3.12 & 3.14 & 3.14 & 3.15 & 3.16 & 3.16 \\
White & 3.37 & 3.38 & 3.4 & 3.4 & 3.41 & 3.41 & 3.43 & 3.43 & 3.44 & 3.45 \\
Hispanic/Latino & 3.11 & 3.06 & 3.12 & 3.1 & 3.1 & 3.1 & 3.13 & 3.12 & 3.13 & 3.14\\
Asian/Pacific Islander & 3.88 & 3.9 & 3.91 & 3.9 & 3.89 & 3.9 & 3.93 & 3.92 & 3.94 & 3.95 \\
American Indian/Alaskan Native & 2.99 & 2.96 & 2.99 & 3.06 & 3.04 & 3.07 & 3.12 & 3.1 & 3.13 & 3.18 \\
Two or more  races/Other & 3.46 & 3.47 & 3.49 & 3.47 & 3.5 & 3.5 & 3.51 & 3.51 & 3.52 & 3.53 \\
\bottomrule
\end{tabular}
\caption{Average scores across Augmented Training data for various demographics. Percentages refer to quantity of original data used within each training set. The remaining instances are synthetic. }\label{demo_scores}
\end{table*}

\begin{table*}
\begin{tabular}{lrrrrrrrrr}
\toprule
& \multicolumn{9}{c}{Percentage of original training data}\\
Score  & 10\% & 20\% & 30\% & 40\% & 50\% & 60\% & 70\% & 80\% & 90\% \\
\midrule
1 & 4.6 & 4.5 & 4.0 & 4.1 & 4.0 & 4.1 & 4.1 & 4.0 & 4.1 \\
2 & 25.3 & 26.1 & 24.3 & 23.8 & 23.2 & 22.5 & 22.1 & 21.8 & 21.3 \\
3 & 30.4 & 30.5 & 31.4 & 31.4 & 31.7 & 31.8 & 31.8 & 31.9 & 31.7 \\
4 & 25.9 & 25.2 & 25.5 & 25.8 & 25.6 & 26.0 & 26.3 & 26.2 & 26.4 \\
5 & 11.5 & 11.1 & 12.0 & 12.0 & 12.6 & 12.6 & 12.8 & 12.8 & 13.1 \\
6 & 2.3 & 2.6 & 2.8 & 2.9 & 2.9 & 3.1 & 3.0 & 3.3 & 3.5 \\ \midrule
w/Aug. Avg. & 3.21 & 3.20 & 3.26 & 3.27 & 3.28 & 3.30 & 3.31 & 3.32 & 3.34\\
Original Avg. & 3.43 & 3.54 & 3.46 & 3.32 & 3.24 & 3.21 & 3.19 & 3.17 & 3.30 \\
\bottomrule
\end{tabular}
\caption{Augmented Training data score-point distribution and average scores across the various percentages.}\label{score_distribution}
\end{table*}


\subsection{Bias}


The reduction of bias is important as it ensures fairness when scoring student essays from different populations and demographics. Using the demographic information provided by the PERSUADE corpus \cite{crossley_persuasive_2022}, we were able to determine whether the synthetic data produced by our Teacher GLM introduced any additional bias or favoritism to our Student models, thus exploring our second RQ. We calculated bias using SMD to determine the difference in essay scores between the synthetic augmented training set and the original training set provided to our Student models. Our comparison also examined the level of bias between each percentage split of original and synthetic data used within the augmented training set compared to the entirety of the original training set. We compared bias in relation to six key demographic variables: gender, race and ethnicity, English Language Learner (ELL) status, disability, and economic status. Figures 3 to \ref{bias_race} depict our findings.

Figure 3 contains four plots illustrating gender, ELL, disability and economic bias between each original/synthetic percentage split of the augmented training set compared to the original training set. Contrary to the average overall positive SMD demonstrated by our Student models, it was found that the synthetic data provided by our Teacher GLM introduced a negative bias when examining the performance of individual demographics. On average, scores across each demographic were lower than those provided by the human annotators of the PERSUADE corpus. Moreover, the more synthetic data incorporated within the augmented dataset then the greater the negative bias (SMD) shown to each demographic. For instance, when only using  10\% to 20\% of the original data, the respective 90\% and 80\% synthetic data resulted in each demographic receiving a far less generous grade, as compared to percentage splits that included more original data. The 10\% original and 90\% synthetic percentage split obtained negative SMDs of -0.116, -0.092, -0.132, and -0.132 for gender, ELL status, disability, and economic status respectively when compared to the original training set. However, the 90\% original and 10\% synthetic percentage split achieved noticeably higher SMDs of -0.014 for gender, -0.010 for ELL status, -0.019 for disability, and -0.018 for economic status, with the increased amount original data leading to a greater likeness between the augmented training set and the original PERSUADE corpus. The Teacher GLM would therefore appear to have a stricter outlook on each student essay when being trained on fewer original samples. It marks each essay on average with a lower grade than PERSUADE's human annotators and thus returns synthetic data with a negative bias in relation to gender, ELL status, disability or economic status. 

The negative bias shown by our Teacher GLM is stronger for demographics that have historically reported worse grades \cite{ormerod_automated_2021}. Students that are English language learners, have a disability, or are at an economic disadvantage are often assigned lower grades than compared to their peers (Table \ref{demo_scores}; Figure 3). Our Teacher GLM marked the essays produced by these students less favorably than PERSUADE's human annotators. For example, English language learners were on average assigned scores with a SMD lower than native English speakers. Likewise, those students with a disability or economic disadvantage were on average provided scores with a greater negative SMD than students without a disability or economic disadvantage respectively. Moreover, our Teacher GLM has also assigned lower scores to students from underrepresented racial backgrounds. For instance, Figure \ref{bias_race} depicts the synthetic data produced by our Teacher GLM as assigning emphatically lower scores for American Indian/Alaskan Native students than compared to their original scoring by an average SMD of -0.106. This may be a result of our Teacher GLM having a more strict adherence to the scoring rubric or penalizing each essay more harshly in regards to small errors in grammar, spelling or general writing.



Table \ref{score_distribution} provides another explanation for our Teacher GLM's lower overall scoring and negative SMDs for each demographic. It was also discovered that our GLM when trained on 10\% of the original data and the remaining synthetic, assigned noticeably less 6 scores (being the highest achievable score) with an average of 2.3 being assigned in the test set than compare to when being trained on 90\% of the original data. This suggests that our Teacher GLM is normalizing its predictions when exposed to a limited amount of original samples, leading to a suppression of higher scores. In turn, this results in a larger number of low scores being assigned to each demographic. 


\section{Discussion}\label{sec:discussion} 

Our experiments show that synthetic data produced by a GLM can be of a similar quality as hand-scored data with the caveat that we need to be very careful about any potential bias introduced by our Teacher GLM as indicated by its varying SMDs. This opens exciting avenues for future research and provides a cost-effective alternative for human hand-scoring. For example, the above findings demonstrate that with the use of synthetic data only 10\% of the PERSUADE corpus's original training set is required to achieve the same state-of-the-art performances on par with that achieved by using the entire dataset, if we can effectively control for SMD variations by other means.

This work has shown that we can use a model-distillation pipeline to effectively reduce the costs of training a model for AES from a QWK point of view. However, there are effects on the SMD that need to be addressed before this pipeline is considered for production \cite{lottridge_psychometric_2023, williamson_framework_2012}. Section \ref{sec:results} shows that the synthetic data produced by our Teacher GLM tends to score each essay lower than compared to the human annotators of the original PERSUADE corpus in regards to particular demographics. Numerous negative biases were demonstrated in relation to gender, disability, ELL and economic status as well as race and ethnicity. Scoring lower grades to each student essay can be problematic since the grade assigned may not accurately reflect the student's ability, progress, or improvement. As such, the SMDs between the predictions of our Student models and the original test set of the PERSUADE corpus need to addressed.



We believe that the lower scores of our synthetic data may be a result of the Reinforcement Learning (RL) techniques used to train the Teacher GLM exacerbating SMDs. One solution to the problem of higher SMDs is to ensure the Student is a regression-based model. In this setting, the model range is a subset of $[0,1]$ which is mapped to a score by applying cutoffs \cite{ormerod_mapping_2022, yang_enhancing_2020}. We can therefore calibrate our Student model to match the expected means of human ratings by imposing a constraint on the average score on the development set. Imposing this constraint on a development set is often conducive to better controlled SMDs on the final validation/test set.

Augmented ``Cyborg Data" is a viable alternative to human annotated datasets. Only 10\% of the original PERSUADE corpus was needed to achieve comparable performance when used in combination with synthetic data. This study provides a method of drastically reducing the cost of human hand-scoring by up to 90\% of its original value. With each student essay costing up to \$5 for a single content expert to mark \cite{EducationGrowthAdvisors2013}, this reduction would amount to a substantial saving in overall expenditure for AES model production. The generation of synthetic data is one example of the capabilities of GLMs in the field of AES. Future research will continue to leverage GLMs and synthetic data for AES and to improve the performance of other tasks related to essay scoring.


\bibliographystyle{plain}
\bibliography{references, custom}

   

\end{multicols}
\end{document}